\documentclass[letterpaper, 10 pt, conference]{ieeeconf}  

\IEEEoverridecommandlockouts                              

\usepackage{graphics} 
\usepackage{amsmath} 
\usepackage{amsmath}
\usepackage{algorithm}
\usepackage[noend]{algpseudocode}
\usepackage{graphicx}
\usepackage{longtable}
\usepackage{graphics}
\usepackage{caption}
\usepackage{subcaption}
\usepackage[normalem]{ulem}
\usepackage{float}
\usepackage{booktabs}
\usepackage{multirow}
\usepackage{siunitx}
\usepackage{amsmath}
\usepackage{hyperref}
\usepackage{float}
\newcommand{\argmin}[1]{\underset{#1}{\operatorname{arg}\,\operatorname{min}}\;}

\newcommand{\RN}[1]{\textup{\uppercase\expandafter{\romannumeral#1}}}

\title{\LARGE \bf 
SROM: Simple Real-time Odometry and Mapping using LiDAR data for Autonomous Vehicles
}

\author{ Nivedita Rufus*$^1$, Unni Krishnan R. Nair*$^1$,
A. V. S. Sai Bhargav Kumar$^{1}$,
Vashist Madiraju$^{1}$, \\ and
K. Madhava Krishna$^{1}$
\thanks{$^{1}$ The authors are with Robotics Research Center (RRC), IIIT-Hyderabad India.* Equal Contribution
{\tt\small nivedita.rufus@research.iiit.ac.in},
{\tt\small unni.krishnan@research.iiit.ac.in},
{\tt\small vseetharam.a@research.iiit.ac.in},
{\tt\small vashist.madiraju@students.iiit.ac.in},
{\tt\small mkrishna@iiit.ac.in}
}
}

\begin{document}

\maketitle
\thispagestyle{empty}
\pagestyle{empty}

\begin{abstract}
In this paper, we present SROM, a novel real-time Simultaneous Localization and Mapping (\textbf{SLAM}) system for autonomous vehicles. The keynote of the paper showcases SROM's ability to maintain localization at low sampling rates or at high linear or angular velocities where most popular LiDAR based localization approaches get degraded fast. We also demonstrate SROM to be computationally efficient and capable of handling high-speed maneuvers. It also achieves low drifts without the need for any other sensors like IMU and/or GPS. Our method has a two-layer structure wherein first, an approximate estimate of the rotation angle and translation parameters are calculated using a Phase Only Correlation (POC) method. Next, we use this estimate as an initialization for a point-to-plane ICP algorithm to obtain fine matching and registration. Another key feature of the proposed algorithm is the removal of dynamic objects before matching the scans. This improves the performance of our system as the dynamic objects can corrupt the matching scheme and derail localization. Our SLAM system can build reliable maps at the same time generating high-quality odometry. We exhaustively evaluated the proposed method in many challenging highways/country/urban sequences from the KITTI dataset and the results demonstrate better accuracy in comparisons to other state-of-the-art methods with reduced computational expense aiding in real-time realizations. We have also integrated our SROM system with our in-house autonomous vehicle and compared it with the state-of-the-art methods like LOAM and LeGO-LOAM.  

\end{abstract}

\section{Introduction}
\begin{figure}[t] 
        \includegraphics[width=0.48\textwidth]{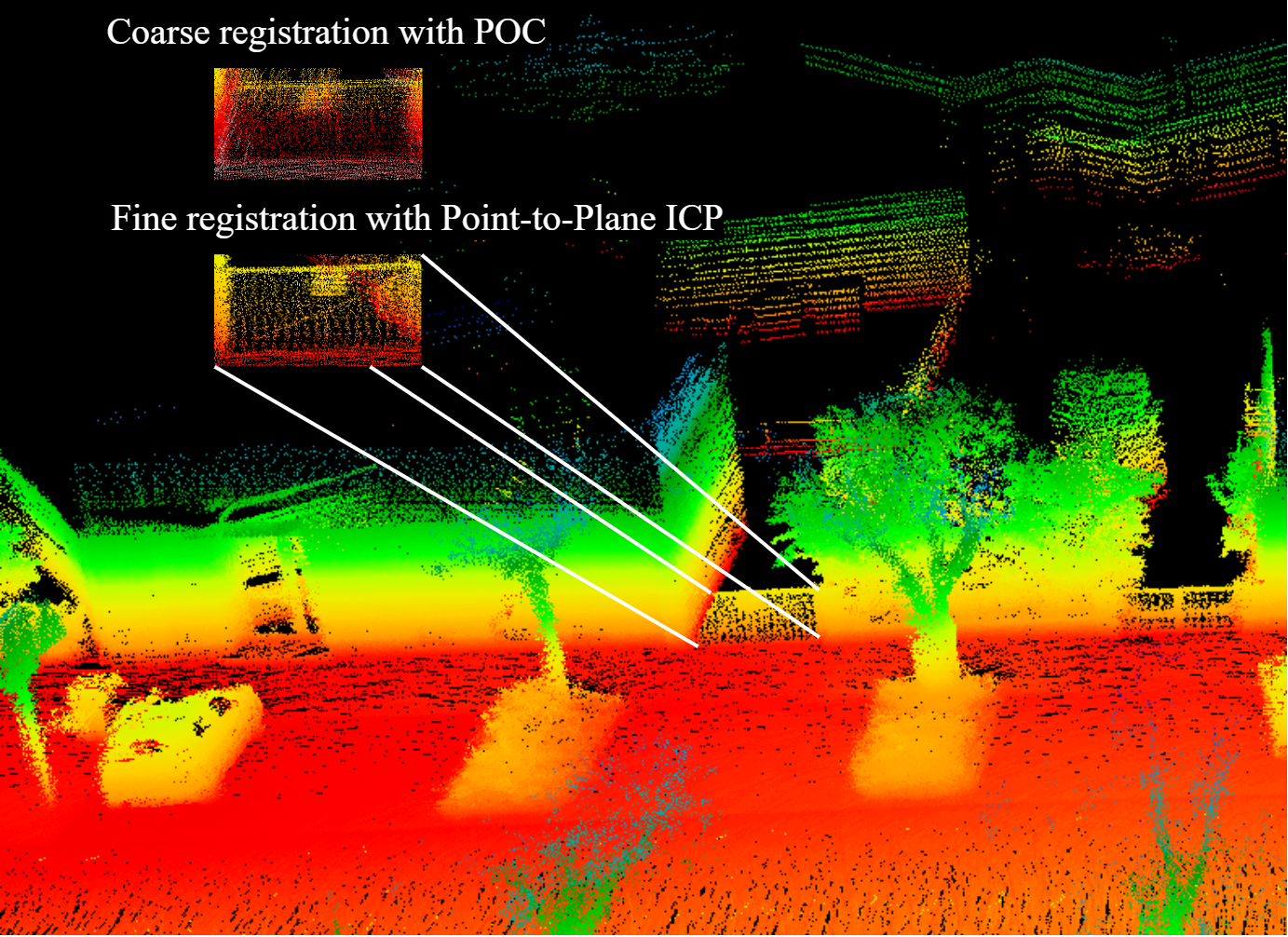}
        \caption{\small We present a 6-DOF odometry and mapping solution for a moving LiDAR. Our proposed method estimates the odometry in two sequential steps. The first step gets a rough odometry estimate (left top corner). This is used as an initialization for the Point-to-Plane ICP which refines the transformation and gives the complete 6-DOF transformation. As a consequence of this, the vertical railings of the gate are accurately registered as shown in the expanded inset.
        \normalsize}
        \vspace{-.5cm}
        \label{teaser}
\end{figure}
\begin{figure*}[t]
\centering
\includegraphics[width=0.9\textwidth]{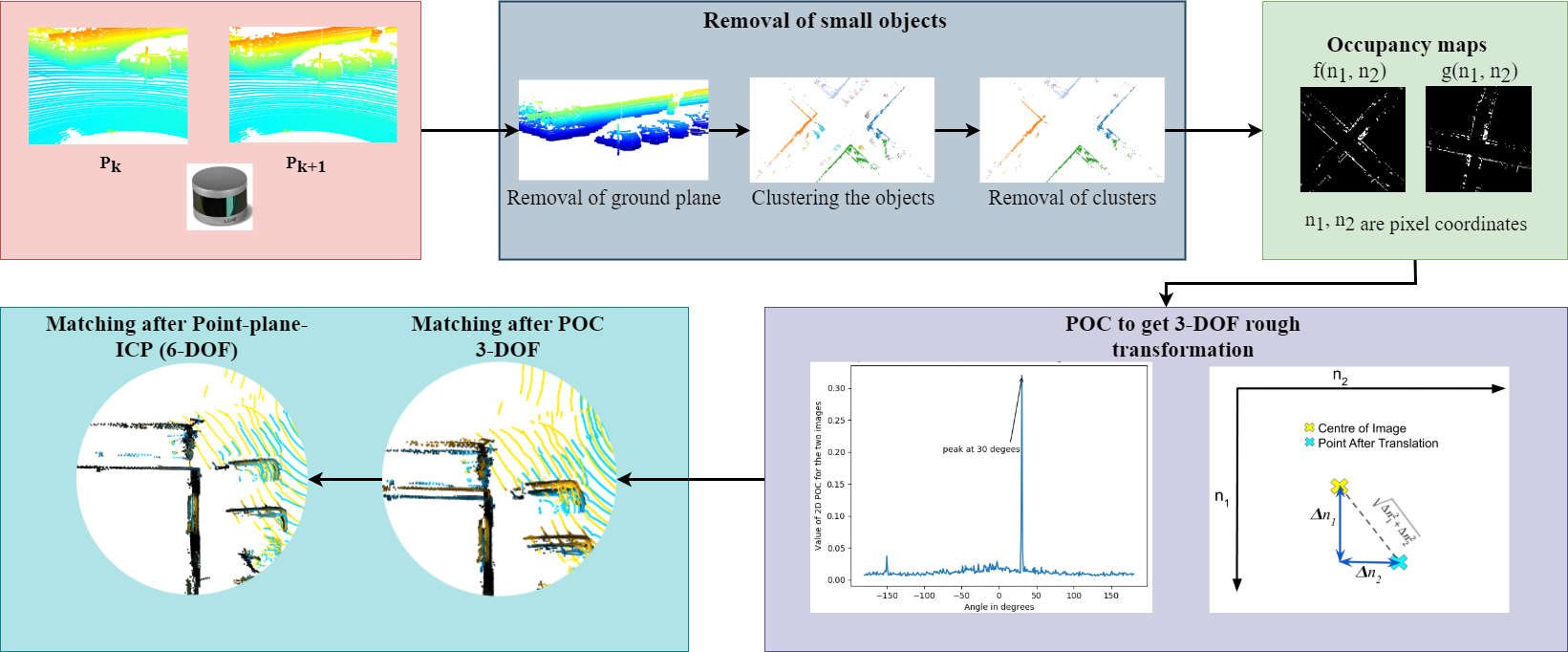}
\caption{\small An overview of the SROM system\normalsize}
\vspace{-.5cm}
\label{pipeline}
\end{figure*}
Simultaneous Localization and Mapping (SLAM) has been an important research problem in the field of autonomous driving \cite{Zhang2014LOAMLO}. All autonomous vehicles need robust SLAM systems, as accurate localization and mapping of the environment surrounding is critical for safe navigation. These systems estimate the motion of the ego vehicle from its perception sensors. There are several SLAM solutions based on different sensors like cameras, depth sensors, radars or a combination of these. But LiDARs hold an advantage over the rest because they are insensitive to the lighting conditions and the optical texture of the environment, therefore, helping in generating consistent maps. Real-time rendering of a 3D map from a stream of large point clouds is a challenging problem considering the rotation of the LiDAR in addition to the ego vehicle's motion. Robust SLAM systems with low computational complexity are needed to calculate the ego-motion of the vehicle without relying on additional sources for localization.

 We present SROM, a SLAM system based on the LiDAR data to provide mapping and localization on the autonomous vehicle in real-time. Further, we demonstrate the robustness of our method to high angular and linear velocities. Our algorithm can accurately register the environment at $10Hz$ and demonstrates superior performance in challenging scenarios where the autonomous vehicle is performing high speed maneuvers (Table \ref{tab:table1}, Fig. \ref{fig:2x}, Fig. \ref{fig:5x} and Fig. \ref{fig:campus}).
 
Our method has a two-layer sequential structure which calculates a rough transformation and fine matching respectively. The first layer uses a Phase-only-correlation (POC) based matching \cite{Nagashima2007AHR} for calculating the approximate estimate of the rotation angle and the translation parameters between two consecutive LiDAR scans. In the second layer, the estimate is further refined by using the output from the previous layer as an initialization for the point-to-plane ICP algorithm. This initialization plays an important role in getting good correspondences between the two scans for the ICP algorithm even at high speeds. This results in high-precision estimates for the ego-motion and the map.

Another key feature of the proposed work is the removal of dynamic objects from the scans before matching. This improves the performance and the accuracy as dynamic objects often corrupt the matching scheme and derail the localization \cite{deschaud2018imls}. The removal of the dynamic object also aids in achieving low drift in odometry without the need for any additional inertial measurement units.

We evaluated our method on the KITTI dataset \cite{articlekitti} and also integrated it with our autonomous vehicle to show its vastly superior performance during high-speed maneuvers. 

\section{Related Work}

\label{sec:relWork}
LiDAR is ubiquitously used in the perception framework of autonomous vehicles \cite{nuchter20076d}. There are several challenges to achieve real-time LiDAR-SLAM. 2D LiDARs can also be used for localization and mapping applications \cite{kohlbrecher2011flexible}. The majority of the 3D LiDAR SLAM approaches are generally variants of the Iterative Closest Point (ICP) scan matching. There are several variants based on ICP from scan-model based registration methods \cite{rusinkiewicz2001efficient} to point-to-plane matching. 
Methods like \cite{pomerleau2013comparing} use standard ICP to match the laser scans from different sweeps. \cite{hong2010vicp} proposes a two-step method comprising of velocity estimation followed by distortion compensation, which helps in reducing the motion distortion if the scan rates are not high. Distortion by single-axis 3D LiDAR is addressed using a similar technique in \cite{moosmann2011velodyne}. But these methods cannot handle the distortions if the scan rate is very low as in the case of 2-axis LiDAR\cite{velas2016collar}. In such cases, the LiDAR scan registration can be done by using the state estimation from IMU along with visual odometry \cite{scherer2012river}. The same can be done by using extended Kalman filters \cite{montemerlo2002fastslam} or particle filter \cite{thrun2002probabilistic}.  

In \cite{pandey2011visually} the ICP algorithm is modified by fusing with RGB cameras that are omnidirectional with the LiDAR scan data. In \cite{dimitrievski2016robust}, the state estimation is reduced from 6 to 3 degrees of freedom. The drawback of this method is that it can not estimate the height differences in the registration. Methods like \cite{deschaud2018imls} provide a scan-to-model approach on the 3D data which results in low global drifts. But the computational complexity of this method doesn't aid in real-time realization. Hence, it cannot be used for real-time applications on autonomous vehicles.

 LOAM \cite{Zhang2014LOAMLO} was able to achieve accurate and precise registrations of the scan data from the LiDAR. This was further improved by fusing the data from RGB-D camera in \cite{zhang2015visual}. Generalized ICP (GICP) was proposed in \cite{segal2009generalized} which replaced the point-to-point matching by plane-to-plane matching. A lightweight ground optimized method is presented in \cite{shan2018lego}. But the methods mentioned above derail when autonomous vehicles perform high-speed maneuvers \cite{kumar2018novel}. To address this we provide a novel system that can provide localization and mapping in real-time which is robust enough to handle high-speed maneuvers. Our method uses a POC based matching \cite{Nagashima2007AHR} for determining a prior estimate. This is then used as an initialization for the point-to-plane ICP algorithm to determine the fine transformation.  

\section{Notations and Problem Description}
The following coordinate frame notations will be used throughout the paper.  Let $P_k$ be the point cloud received from the LiDAR at sweep $k$.
\begin{itemize}
    \item The LiDAR frame is represented by \{$L$\}, where the $x$-axis points to the left, $y$-axis points upwards, $z$-axis points forward. Also the point $i$ in \{$L$\} where $i$ $\epsilon$ $P_k$ is represented as $X^L(k,i)$.
  \item The World frame is represented by \{$W$\}, which coincides with \{$L$\} at $k = 1$ and point $i$ in \{$W$\} where $i$ $\epsilon$ $P_k$ is represented as $X^W(k,i)$.
\end{itemize}
\normalsize
We define our problem statement as, given a sequence of LiDAR sweeps $P_1,P_2,..., P_k$, $k$ $\epsilon$ $Z^+$, estimate the ego vehicles pose as $T^{W}_{L}(k+1)$ for each sweep $k+1$ where $T^{W}_{L}$ represents the relative pose of $\{L\}$ with respect to $\{W\}$ and build a map as a set of points $M$.

\section{Overview of the system}
\label{sec:overview}
 An overview of the proposed software pipeline is shown in Fig. \ref{pipeline}.  We use a RANSAC based iterative plane fitting approach for ground plane estimation. The point cloud is rectified by performing relative inverse rotation to the ground normal estimated. The rectified point cloud is then projected on to the world plane and the prior correspondences are generated using 2D-Phase Only Correlation (POC). This output is further processed by iterative point-to-plane ICP algorithm to generate the fine correspondences which register and map the undistorted point cloud at the frequency of 10Hz. We present this method in the following sections.

\section{Pose Estimation}

\subsection{Ground Plane Estimation}
We start with the estimation of the ground plane from the LiDAR point cloud $P_k$. This is done by using a RANSAC based algorithm \cite{dimitrievski2016robust}. We take a random subset of points from the LiDAR point cloud to fit a plane equation. Using this equation, we calculate the average distance to all the other points from the plane. This process is repeated and until the subset with the lowest mean distance is selected and a new plane is fitted using the least-squares method to get the optimal ground plane. We get a robust ground plane with minimal iterations, by limiting the set of points that RANSAC chooses from. 

\subsection{Dynamic Object Set Removal}
\label{DOSR}
\begin{figure}[htb]
    \includegraphics[width=0.48\textwidth]{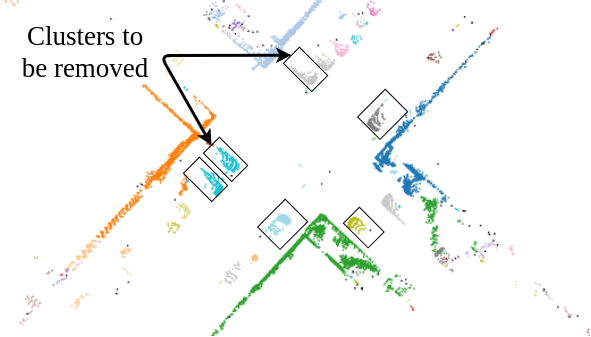}
    \caption{\small This figure shows the clustering and removal of small objects from the point cloud using the DBSCAN algorithm.
    \normalsize}
    \label{fig:outlier_removal}
\end{figure}

We remove all the dynamic objects from the scan before matching to achieve low drifts \cite{deschaud2018imls}. This requires semantic information of the scan which increases the computational complexity. Instead, we perform small object removal, which achieves a similar result. This is done by first removing the points corresponding to the ground plane. Then we cluster the remaining of the point cloud using the DBSCAN algorithm \cite{ester1996density}. We remove the clusters having a size less than a threshold that can be tuned. After this, we add back the points corresponding to the ground plane. We use a density parameter of 0.5 for the clustering and all the clusters having a bounding box lesser than $10m, 10m, 4m$ (these values were found to work well for the KITTI dataset) in $z, x, y$ respectively in \{$L$\}. Fig. \ref{fig:outlier_removal} shows our small object removal strategy on a scanned point cloud. After the removal of these objects, the point cloud is rectified by performing inverse rotation with respect to the estimated ground plane.

\subsection{POC based Scan Matching}
\label{sec:FFT}
We perform a POC (Phase Only Correlation) based scan matching on the rectified point cloud. We define the occupancy map as a 2D grid $m_k$ with grid elements $m_k(n_1,n_2)$ where $n_1$ and $n_2$ are pixel coordinates. $P_1,P_2...P_k$ are scans obtained from the LiDAR till the $k^{th}$ sweep. Let $f(n_1, n_2)$, $g(n_1, n_2)$ be the local occupancy maps at $k^{th}$, ${k+1}^{th}$ sweep respectively, where $n_1 = -\frac{(N-1)}{2},...,\frac{(N-1)}{2}$, $n_2 = -\frac{(N-1)}{2},...,\frac{(N-1)}{2}$, for an $N \times N$ image \cite{dimitrievski2016robust}. This is shown in (\ref{eqn:occ}) where $p(m_k({n_1,n_2}) | P_1,...,P_k)$ is the probability of a cell being occupied. We transform the scan matching problem to an image registration problem.


\vspace{-.2cm}
\small
\begin{subequations}
\label{eqn:occ}
\begin{equation}
p(m_k({n_1,n_2}) | P_1,...,P_k)=1-\frac{1}{e^{l({n_1,n_2})}}
\end{equation}
\vspace{-.2cm}
\begin{equation}
\begin{aligned}
l({n_1,n_2})&=\log \frac{p\left(m_k({n_1,n_2}) | P_{k}\right)}{1-p\left(m_k({n_1,n_2}) | P_{k}\right)} 
+\log \frac{1-p\left(m_k({n_1,n_2})\right)}{p\left(m_k({n_1,n_2})\right)} \\
&+l_{\text {past}}({n_1,n_2})
\end{aligned}
\end{equation}
\vspace{-.4cm}
\begin{equation}
\begin{split}
f(n_1,n_2) &= m_k(n_1,n_2)\\
g(n_1,n_2) &= m_{k+1}(n_1,n_2)
\end{split}
\end{equation}
\end{subequations}
\normalsize


The POC estimates the translational shift between two images. To estimate the rotational shift, we map the amplitude spectra to its polar space and perform POC \cite{Nagashima2007AHR}.

\begin{figure}[htb]
    \begin{subfigure}[b]{0.24\textwidth}
          \includegraphics[width=0.9\linewidth]{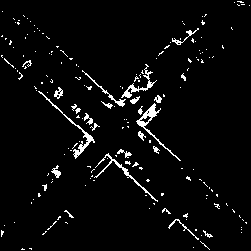}
            \caption{\small $f(n_1, n_2)$ \normalsize}
            \label{occ_f}
    \end{subfigure}
    \begin{subfigure}[b]{0.24\textwidth}
          \includegraphics[width=0.9\linewidth]{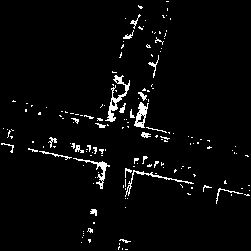}
            \caption{\small $g(n_1,n_2) \normalsize$}
            \label{occ_g}
    \end{subfigure}
    \caption{\small (a) and (b) represent the local occupancy maps for the estimation of the 3-DOF transformation using POC. \normalsize}
\vspace{-.2cm}
    \label{fig:occ_map}
\end{figure}

First, the discrete scan  $f(n_1, n_2)$ and $g(n_1, n_2)$ (Fig. \ref{fig:occ_map}) are subjected to a 2D Hanning window in order to reduce the discontinuity at the corners. Then the amplitude spectra is obtained by calculating the FFT using the (\ref{eqn:1})

\small
\begin{subequations}\label{eqn:1}
    \begin{equation} 
    |F(u,v)| = |\mathcal{F}(f(n_1, n_2))|
    \end{equation}
\vspace{-.4cm}
    \begin{equation} 
    |G(u,v)|  = |\mathcal{F}(g(n_1, n_2))|
    \end{equation}
\end{subequations}
\normalsize
\vspace{-.4cm}

where, $\mathcal{F}(.)$ denotes the Fourier transform and $F(u,v),G(u,v)$  are the amplitude spectra having their non-zero frequency component in the centre of the spectrum. $u$, $v$ are the pixel coordinates in the Fourier space and ranges from $-\frac{(N-1)}{2},...,\frac{(N-1)}{2}$, for an $N \times N$ image, 

The FFT relation between the scans can be given by (\ref{eqn:poc})
\vspace{-0.1cm}
\small
\begin{equation}
    g(n_1,n_2) = f(n_1-\Delta n_1,n_2-\Delta n_2)
    \label{eqn:poc}
\end{equation}
\normalsize

where $\Delta n_1, \Delta n_2$ are the translation along $n_1, n_2$ directions respectively and its Fourier transform is given by (\ref{eqn:fft}) 


\small
\begin{equation}
\label{eqn:fft}
    G(u,v) = F(u,v)e^{-2\pi i(\frac{u\Delta n_1}{N}+\frac{v\Delta n_2}{N})}
\end{equation}
\normalsize



\begin{figure}[htb]
\centering
    \begin{subfigure}[b]{0.48\textwidth}
    \centering
          \includegraphics[width=0.8\linewidth]{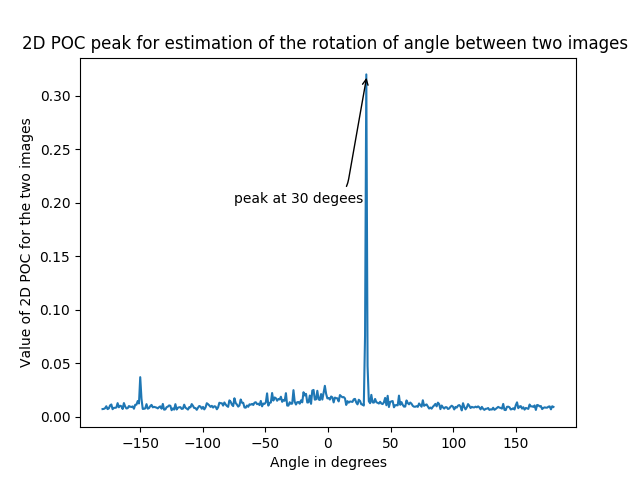}
          \vspace{-.2cm}
            \caption{\small The result of POC on the polar maps of Fig. \ref{occ_f} and Fig. \ref{occ_g} to estimate the angle of rotation.\normalsize}
            \vspace{-.1cm}
            \label{fig:peak:a}
    \end{subfigure}
    \begin{subfigure}[b]{0.48\textwidth}
    \centering
          \includegraphics[width=0.8\linewidth]{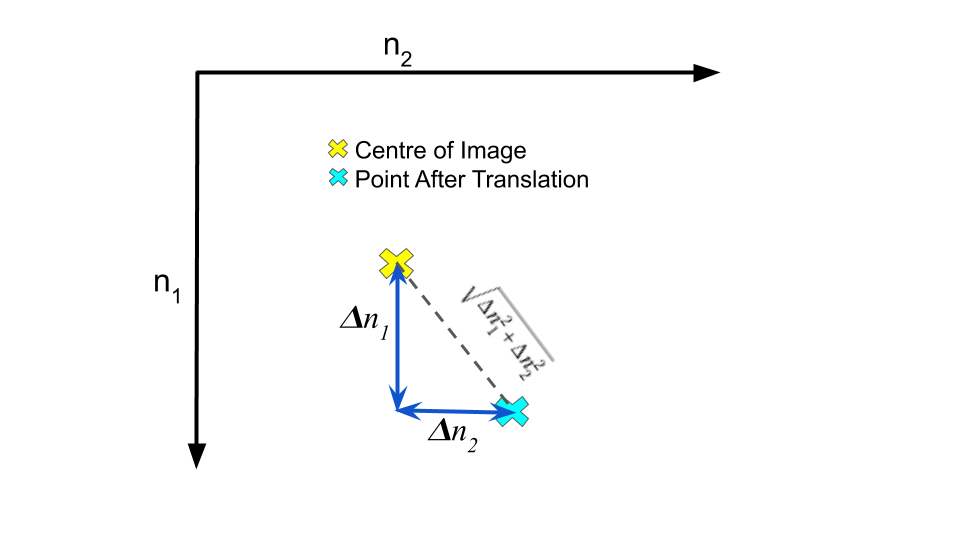}
          \vspace{-0.4cm}
            \caption{\small The result of POC to estimate the translation.\normalsize}
            \vspace{-0.2cm}
            \label{fig:peak:b}
    \end{subfigure}
    \caption{\small Results of the POC to find angle of rotation and translation between Fig. \ref{occ_f} and Fig. \ref{occ_g}. \normalsize}
    \vspace{-.6cm}
\end{figure}

We then calculate the cross-power spectrum of the two images from their Hadamard product to obtain the phase difference given by (\ref{eqn:cross-power2}),
\small
\begin{subequations}
\begin{equation}
\label{eqn:cross-power1}
\begin{split}
     R(u,v) &= \frac{F(u,v) G^\ast(u,v)}{|F(u,v) G^\ast(u,v)|} \\
     &= \frac{F(u,v)G^\ast(u,v) e^{-2\pi i(\frac{u\Delta n_1}{N}+\frac{v\Delta n_2}{N})}}{|F(u,v)G^\ast(u,v)|}
 \end{split}
 \end{equation}
 \begin{equation}
\label{eqn:cross-power2}
     R(u,v) = e^{-2\pi i(\frac{u\Delta n_1}{N}+\frac{v\Delta n_2}{N})}
   \end{equation}
\end{subequations}
\normalsize
where, $G^\ast(u,v)$ denotes the complex conjugate of $G(u,v)$.
On taking the inverse of (\ref{eqn:cross-power2}) we get a Kronecker delta function which results in a peak giving the translation in $n_1$ and $n_2$ using (\ref{inverse}) which is further illustrated in Fig. \ref{fig:peak:b}.
\small
\begin{equation}
\label{inverse}
    r(n_1,n_2) = \delta(n_1+\Delta n_1, n_2+\Delta n_2)
\end{equation}
\normalsize
To estimate the rotation, we map the amplitude ($F(u,v), G(u,v)$) spectra to the polar space as 
$F_p(l_1, l_2)$ and $G_p(l_1,l_2)$ , where $l_1$, $l_2$ are the pixel coordinates of the polar mapped image and range from $-\frac{(N-1)}{2},...,\frac{(N-1)}{2}$. This converts the angular shift to a translational shift. Then using POC we compute the shift $\Delta l_2$ in $l_2$. The rotation angle is then given by (\ref{eqn: theta}) and is illustrated in Fig. \ref{fig:peak:a},
\vspace{-0.2cm}
\small
\begin{equation}
\label{eqn: theta}
    \theta = \pi\frac{ \Delta l_2}{N} 
\end{equation}
\normalsize
Finally COG (Centre Of Gravity) function fitting is applied  to ensure sub-pixel accuracy resulting in the 3-DOF transformation $T'(k+1)$ given by (\ref{eqn:T_init}),
\small
\begin{equation}\label{eqn:T_init}
    T'(k+1) = \begin{bmatrix}
    \cos{\theta}& -\sin{\theta}& 0& \Delta n_1\\
    \sin{\theta}& \cos{\theta}& 0& \Delta n_2\\
    0& 0& 1& 0\\
    0& 0& 0& 1
    \end{bmatrix}
\end{equation}
\normalsize
$T'(k+1)$ is used as an initialization for the point-to-plane ICP discussed in section(\ref{subsec:icp}) for getting precise registration with significantly reduced number of iterations to achieve convergence.
\subsection{Point-to-Plane ICP}
\label{subsec:icp}
The estimates from the POC are used as an initialization to the Point-to-Plane ICP (Iterative Closest Point) algorithm \cite{rusinkiewicz2001efficient} to get a 3D rigid-body transformations $T$ between the two point clouds $P_k$, $P'_{k+1}$ such that $T$ transforms $P'_{k+1}$ to minimise the total error between corresponding points with a chosen error metric, i.e. we calculate $T_{icp}$ from (\ref{eqn:topt}),

\small
\begin{equation}\label{eqn:topt}
T_{icp} = \argmin{T} \sum_i ((T\cdot s_i - d_i)\cdot n_i)^2
\end{equation}
\vspace{-.4cm}
\begin{align*}
    &s_i = \begin{bmatrix}
    P'_{k+1_x}& P'_{k+1_y}& P'_{k+1_z}& 1
    \end{bmatrix}^{T}\\
    &d_i =\begin{bmatrix}
     P_{{k}_x}& P_{{k}_y}& P_{{k}_z}& 1
    \end{bmatrix}^{T}\\
    &n_i = \begin{bmatrix}
    n_{i_x}& n_{i_y}& n_{i_z}& 0
    \end{bmatrix}^{T}\\
     &P'_{k+1} = T'(k+1)\times P_{k+1}
\end{align*}
\normalsize
where $n_i$ is the unit normal vector at $d_i$. The 6-DOF transformation matrix $T_{icp}$ comprises of a rotation matrix $R_{icp}(\alpha,\beta,\gamma)$ and a translation matrix $t_{icp}(t_x,t_y,t_z)$, i.e.,
\vspace{-0.1cm}
\small
\begin{equation}
    T_{icp} = t_{icp}(t_x,t_y,t_z)\cdot R_{icp}(\alpha,\beta,\gamma)
\end{equation}
\normalsize
\vspace{-0.1cm}
where, $\alpha,\beta$ and $\gamma$ are the rotations and $t_x,t_y,t_z$ are translations about \textit{x}-axis, \textit{y}-axis and \textit{z}-axis respectively.

\subsection{Final Map Generation}


\begin{figure*}[htb]
    \centering
    \begin{subfigure}[b]{0.3\textwidth}
        \includegraphics[width=\textwidth, height=5cm]{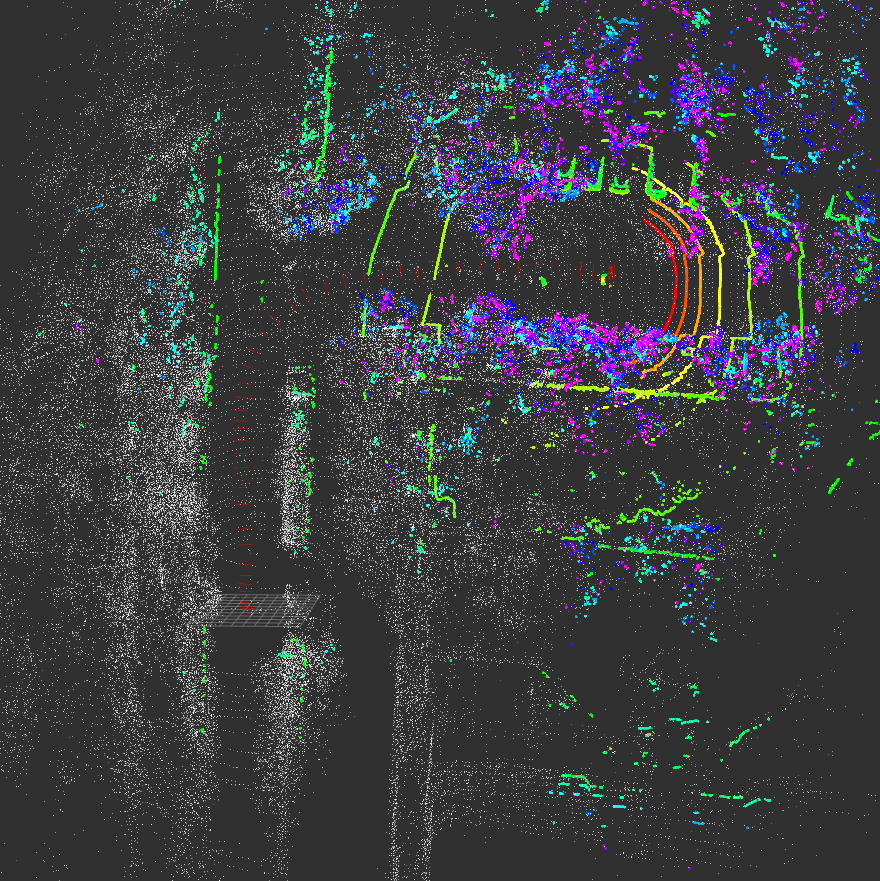}
        \caption{\small LOAM \normalsize}
        \label{fig:loam2}
    \end{subfigure}
    \begin{subfigure}[b]{0.3\textwidth}
        \includegraphics[width=\textwidth,height=5cm]{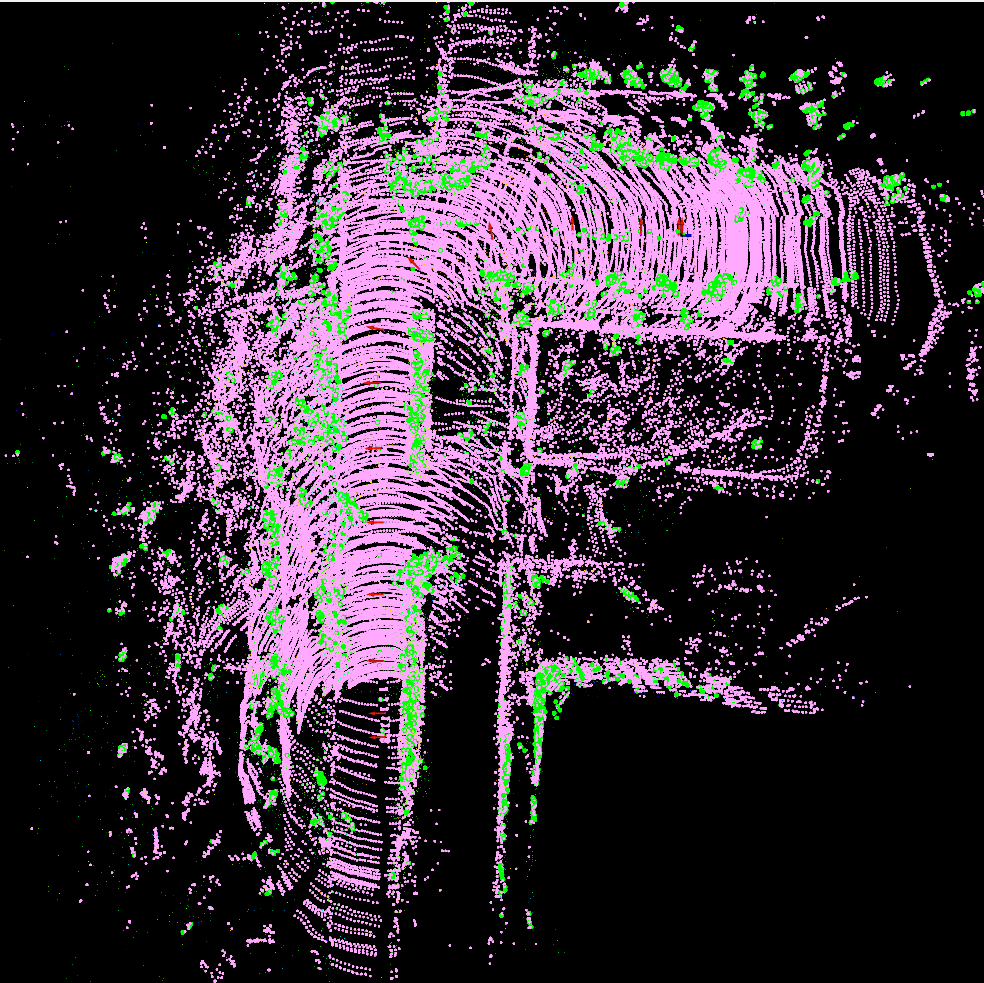}
        \caption{\small LeGO-LOAM \normalsize}
        \label{fig:lego2}
    \end{subfigure}
    \begin{subfigure}[b]{0.3\textwidth}
        \includegraphics[width=\textwidth,height=5cm]{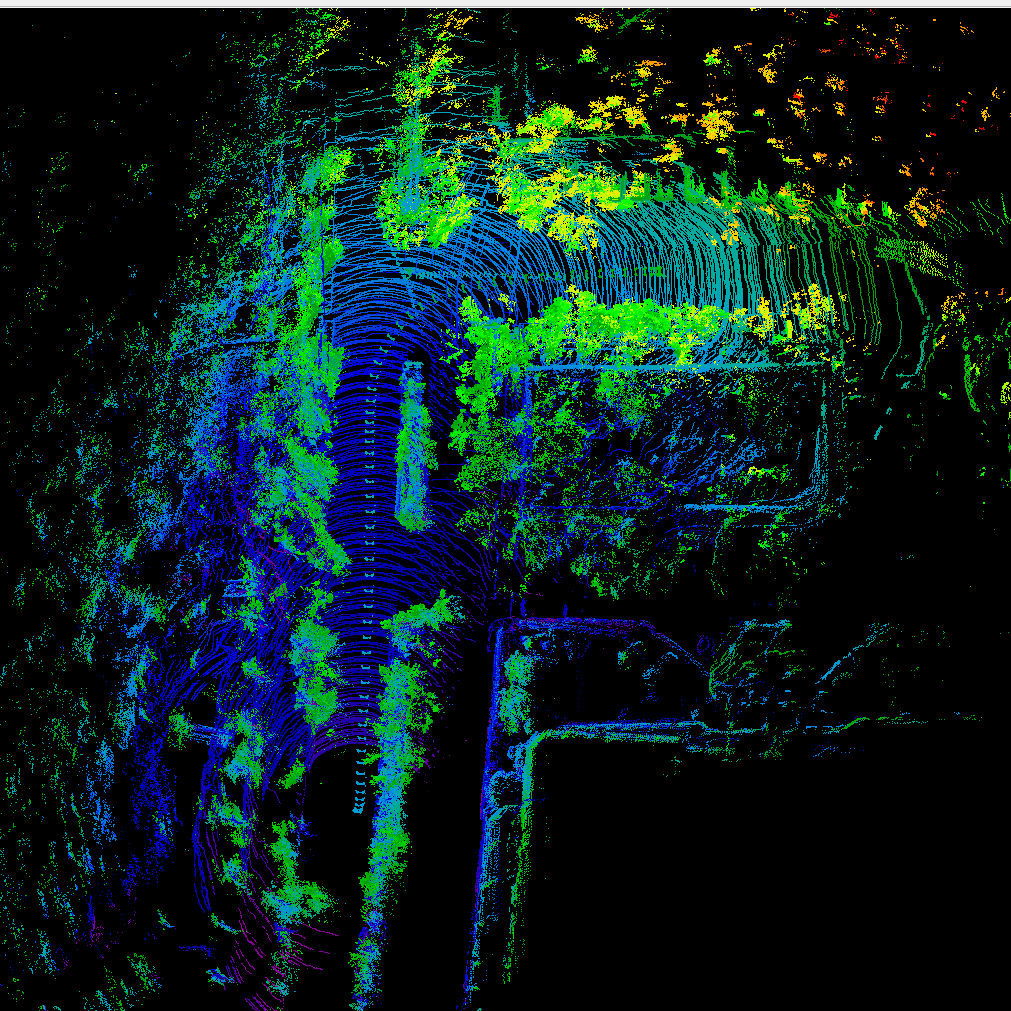}
        \caption{\small SROM (ours)\normalsize}
        \label{fig:srom}
    \end{subfigure}
    \vspace{-0.2cm}
\caption{\small This figure shows the registration of the map when the ego-motion was at twice base speed($40km/hr$) while executing a T-merge maneuver. Though the difference in the qualitative performance is not obvious, we can observe that LOAM and LeGO-LOAM suffer from significant degradation from the values in Table \ref{tab:table1}. \normalsize}
\vspace{-0.2cm}
     \label{fig:2x}
\end{figure*}
\begin{figure*}[htb]
    \centering
    \begin{subfigure}[b]{0.3\textwidth}
        \includegraphics[width=\textwidth, height=5cm]{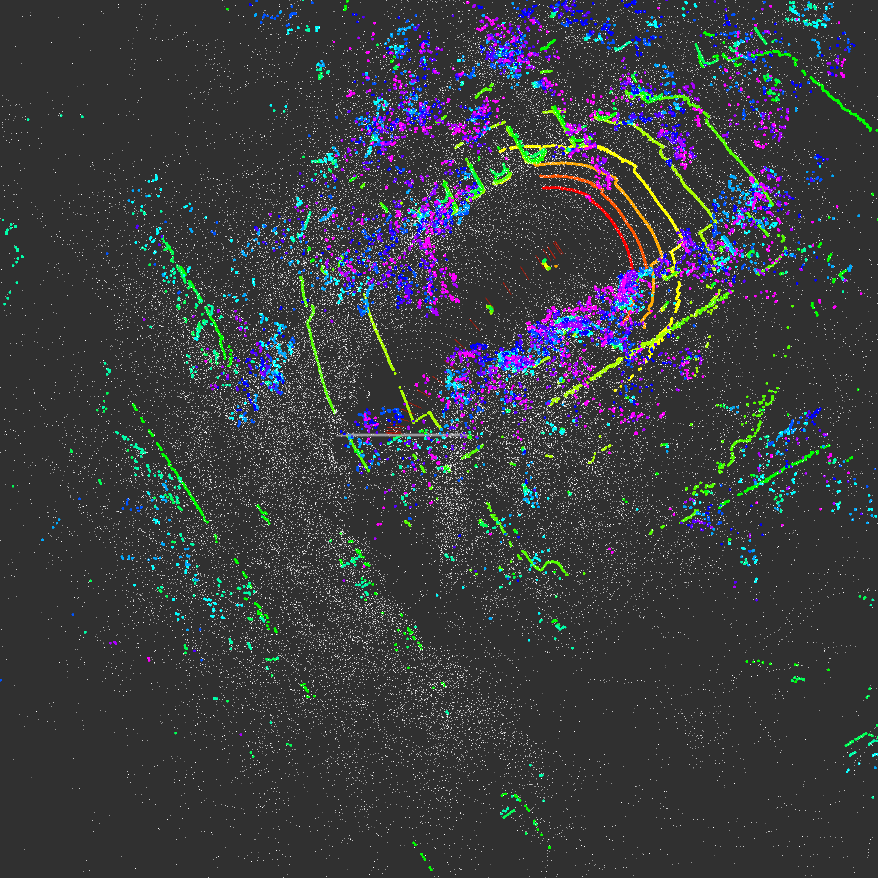}
        \caption{\small LOAM \normalsize}
        \label{fig:loam5}
    \end{subfigure}
    \begin{subfigure}[b]{0.3\textwidth}
        \includegraphics[width=\textwidth,height=5cm]{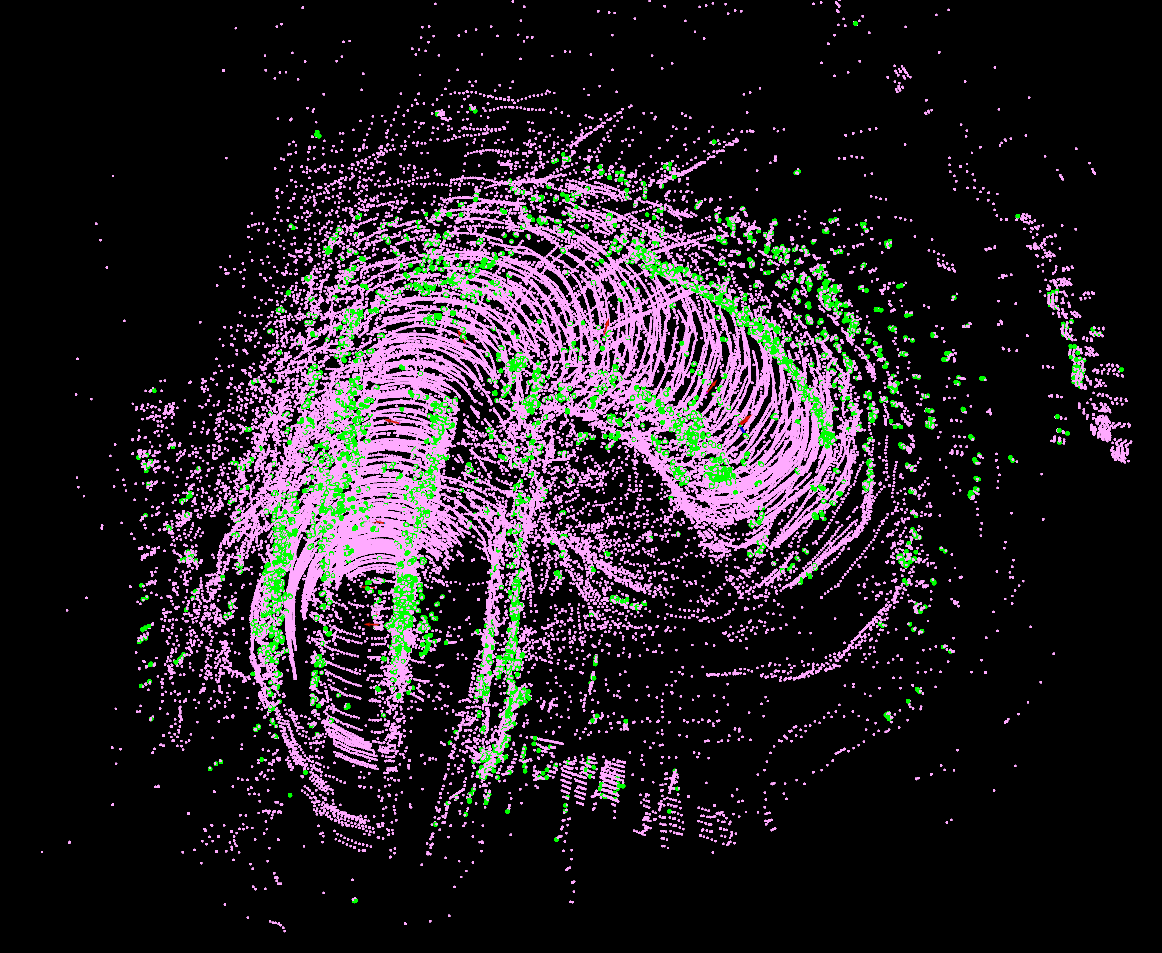}
        \caption{\small LeGO-LOAM \normalsize}
        \label{fig:lego5}
    \end{subfigure}
    \begin{subfigure}[b]{0.3\textwidth}
        \includegraphics[width=\textwidth,height=5cm]{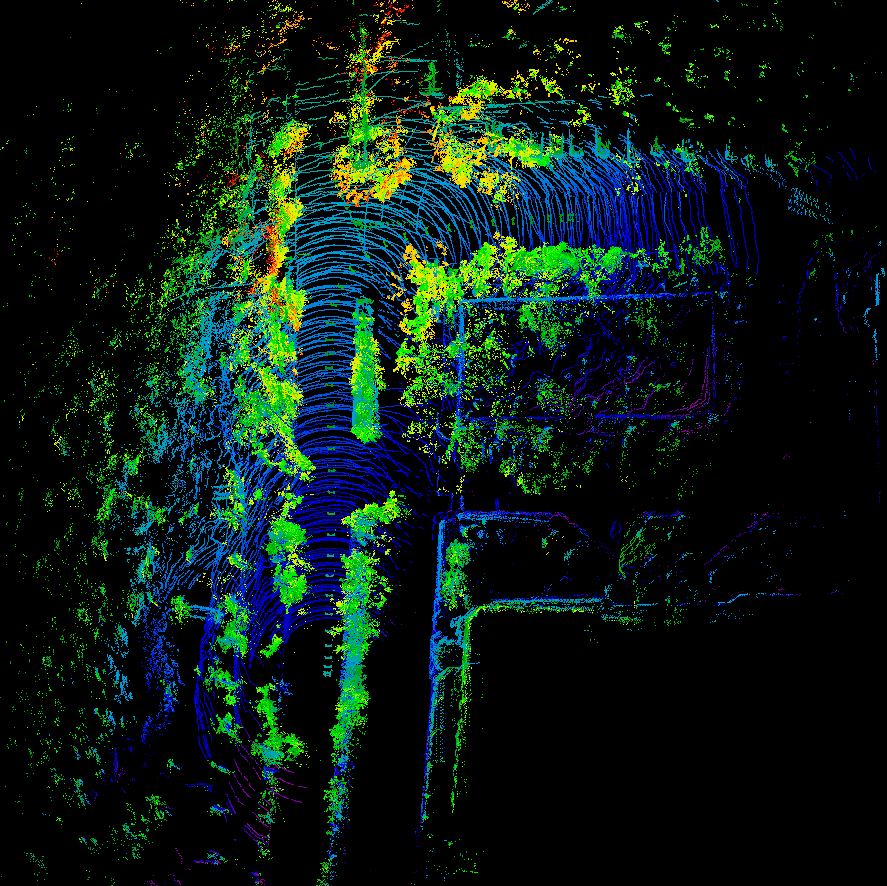}
        \caption{\small SROM (ours)\normalsize}
        \label{fig:srom5}
    \end{subfigure}
\vspace{-0.2cm}
     \caption{\small This figure shows the registration of the map when the ego motion was at five times the base speed($100km/hr$) while executing a T-merge maneuver. It can be observed the that both LOAM and LeGO-LOAM completely lost tracking whereas SROM was still able to give us good odometry estimates with very less degradation Table \ref{tab:table1}. \normalsize}
\vspace{-.5cm}

     \label{fig:5x}
\end{figure*}
The pose at the kth sweep is calculated as
\vspace{-0.2cm}
\small
\begin{equation}
T^{W}_{L}(k+1) = \prod^{k+1}_{j=1} (T_{icp}(j) \times T'(j))
\end{equation}
\normalsize
The final map is generated by (\ref{eqn:map}) where $\forall$ $i$ $\epsilon$ $P_{k+1}$.
\small
\begin{subequations}
\begin{equation}
     M = \{\{ {X^L(1,i)} \} \cup \{X^W(2,i)\}...\cup \{X^W(k+1,i)\}\}\\
\end{equation}
\begin{equation}
X^W(k+1,i) = {T^{W}_{L}(k+1) \times X^L(k+1,i)}
\end{equation}
\label{eqn:map}
\end{subequations}
\normalsize
\vspace{-0.5cm}
\section{Experimental Results} \label{results}
\subsection{Simulation Framework and Computational time}
\begin{figure}[htb]
\begin{subfigure}[b]{0.24\textwidth}
        \includegraphics[width=.9\linewidth]{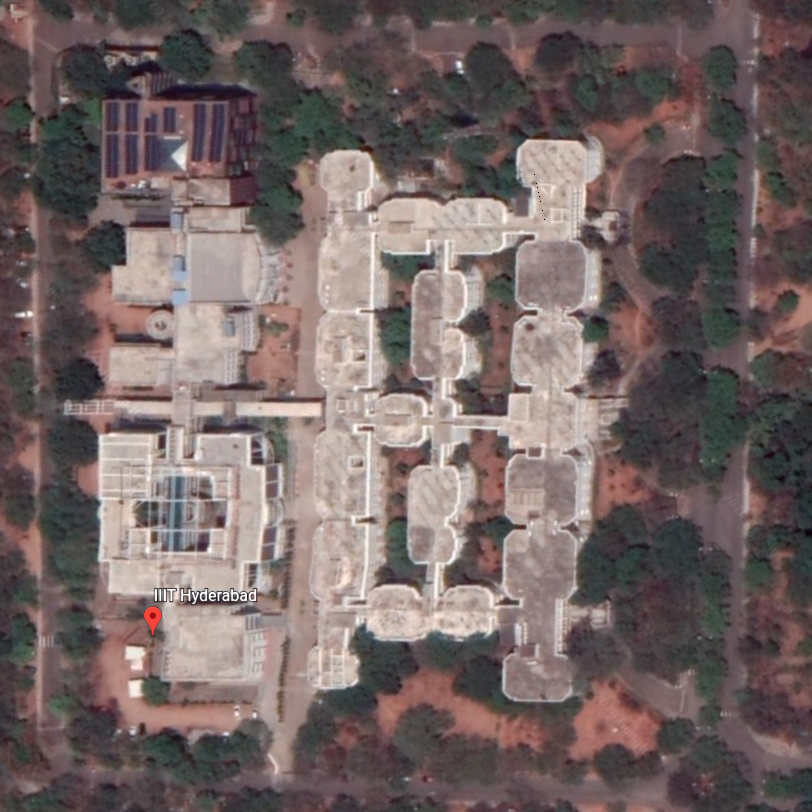}
        \caption{\small Satellite view of our campus\normalsize}
        \vspace{-0.2cm}
        \label{fig:map}
    \end{subfigure}
    \begin{subfigure}[b]{0.24\textwidth}
        \includegraphics[width=.9\linewidth]{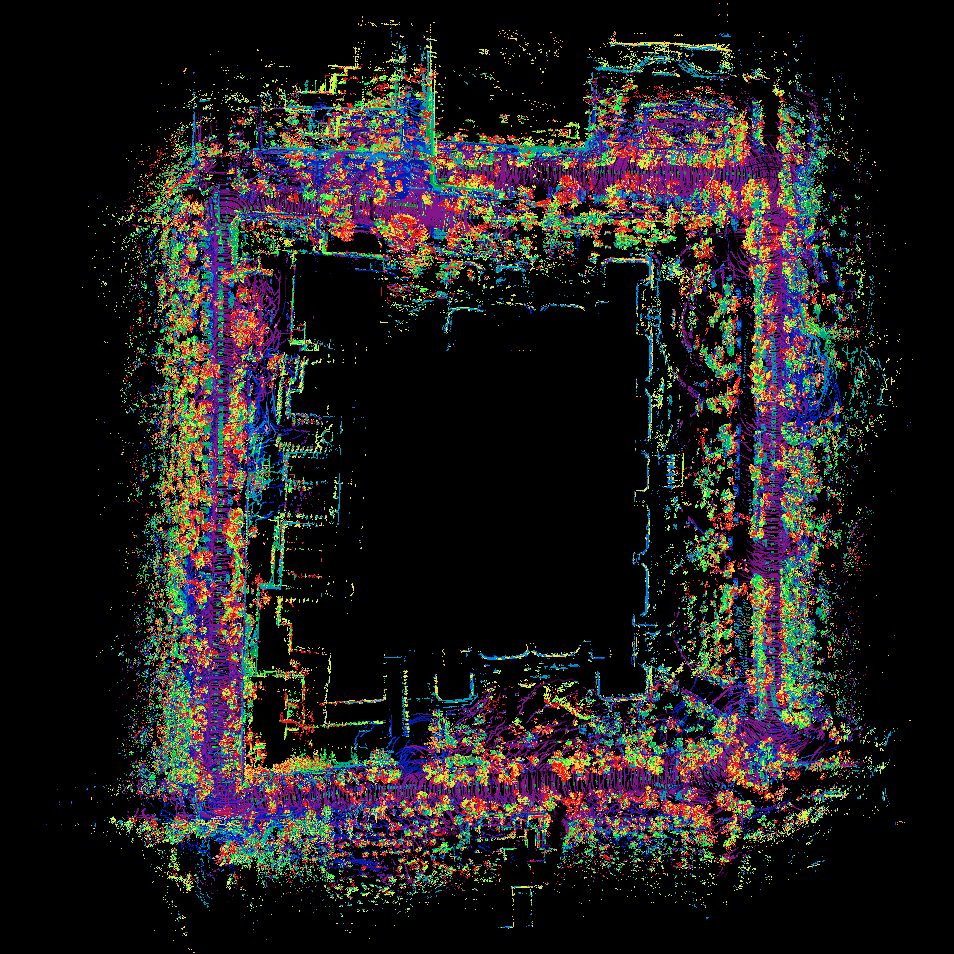}
        \caption{\small Map of our campus (SROM)\normalsize}
        \vspace{-0.2cm}
        \label{fig:sat}
    \end{subfigure}
     \caption{\small Real-time execution of SROM in our campus\normalsize}
     \vspace{-0.4cm}
     \label{fig:campus}
\end{figure}

To evaluate the performance of the proposed framework, we generated the maps for various sequences from KITTI and our campus(done real-time using VLP16, Fig. \ref{fig:campus}). The 3D data processing is done using the Open3D library \cite{Zhou2018}. We also compared our performance to the current state-of-the-art algorithms namely LOAM, LeGO-LOAM in scenarios with different speeds. We simulated the high-speed scenarios by skipping data frames obtained from the LiDAR for the KITTI dataset as shown in Table \ref{tab:table1} and Fig. \ref{fig:2x}, Fig. \ref{fig:5x}. We also evaluated the performance of our approach using two popular sensors HDL64 and VLP16. These two sensors represent a wide range of sensor performance and resolutions that modern LiDAR solutions offer, hence we show that our methods are robust enough to be used in a multitude of scenarios. The simulations were performed on an Intel i7 7700HQ processor @ 2.8 GHz clock speed with 8Gb of ram, and 512 Gb of NVME storage. The average execution time for each cycle including the ICP is around 100ms and all the experimental results shown in this paper are executed at 10Hz.\\
Table \ref{tab:table1} present the \% error per 100m generated by LOAM, LeGO-LOAM and our SROM(with and without dynamic set removal) both on KITTI and real-time implementation. It can be observed that the LOAM and LeGO-LOAM lost tracking(LT) at high speed whereas SROM was able to perform with a low \% error per 100m. Also, the real-time implementation of SROM was done by driving the autonomous car in equipped with VLP-16 a complete loop around the campus. The error generated is  of $0.61\% per 100m$ with and with removing small objects. Fig.\ref{fig:campus} shows the map was successfully generated Owing to the quality of the odometry, we were able to execute high-speed maneuvers autonomously. The following video shows our framework and the dataset evaluations in action. Link:  \href{https://youtu.be/ccTYdJNIzQQ}{https://youtu.be/ccTYdJNIzQQ}.\\
Fig.\ref{fig:2x} and Fig.\ref{fig:5x} demonstrate the performance of LOAM, LeGO-LOAM and SROM while performing a T-merge maneuver at different speeds. It is observed that LOAM and LeGO-LOAM get derailed when the maneuvers are performed at high speeds, whereas SROM was still able to give us good odometry estimates-with very less degradation from  when it was moving at lower speeds. This illustrates the proposed system's robustness to high angular and linear velocities.
\subsection{Downstream Application}

In autonomous driving, it is very vital to have consistent and continuous odometry. Most of the traditional SLAM systems which we tested needs a smooth ego-motion for accurate odometry estimation. We tested a sharp 90-degree turn maneuver at different speeds across an intersection in our university campus. At low speeds, all the SLAM systems perform reasonably well, but we find that LOAM and LeGO-LOAM loose tracking at around five times the base speed. Our system is still able to give a very good estimate of the odometry. Also, we were able to reuse the occupancy maps generated by SROM for path planning.
\vspace{-.4cm}
\begin{center}
\small
\begin{table}[htb]
\begin{tabular}{|c|c|c|c|c|c|}
\hline
\multicolumn{1}{|c|}{\multirow{5}{*}{Seq}} & \multicolumn{1}{c|}{\multirow{5}{*}{Speed}} & \multicolumn{4}{c|}{\%Error per 100m}\\ \cline{3-6}
  &                             &\multirow{4}{*}{LOAM}        &\multirow{4}{*}{LeGO-LOAM}      &\multicolumn{2}{c|}{SROM (Ours)}\\ \cline{5-6}
  
  & & & &with &without \\
  & & & &small &small\\
  & & & &objects &objects\\\hline
   \multirow{3}{*}{01}   &{1X}     &1.43          &1.08               &0.84       &0.84 \\ \cline{2-6}
                         &{2X}     &2.4             &2.3               &0.99       &0.98 \\ \cline{2-6}
                         &{5X}     &LT             &LT               &1.01       &0.98 \\ \hline
   \multirow{3}{*}{02}   &{1X}    &0.92          &0.81               &0.79       &0.78 \\ \cline{2-6}
                         &{2X}     &1.87          &1.13               &0.85       &0.82 \\ \cline{2-6}
                         &{5X}     &LT             &LT               &0.86       &0.86\\\hline
   \multirow{3}{*}{03}   &{1X}     &0.86          &0.99               &0.67       &0.68 \\ \cline{2-6}
                         &{2X}     &0.89             &1.06               &0.69       &0.66 \\ \cline{2-6}
                         &{5X}     &LT             &LT               &0.73       &0.72 \\ \hline
   \multirow{3}{*}{04}   &{1X}     &0.71          &0.69               &0.41       &0.39 \\ \cline{2-6}
                         &{2X}     &0.88          &0.97               &0.49       &0.48 \\ \cline{2-6}
                         &{5X}     &LT             &LT               &0.61       &0.61 \\ \hline
    \multirow{3}{*}{05}  &{1X}     &0.57          &0.68               &0.69       &0.67 \\ \cline{2-6}
                         &{2X}     &0.87          &0.99               &0.71       &0.70 \\ \cline{2-6}
                         &{5X}     &LT             &LT               &0.87       &0.86 \\ \hline
  \multirow{3}{*}{IIITH}   &{1X}     &0.41        &0.74               &0.47       &0.44 \\ \cline{2-6}
                          &{2X}     &0.57         &0.93             &0.55       &0.55 \\ \cline{2-6}
                          &{5X}     &LT             &LT               &0.61       &0.61 \\ \hline
\end{tabular} 
\caption{\small The \% error per $100m$ generated by LOAM (results are taken from \cite{Zhang2014LOAMLO}), LeGO-LOAM (results are taken from \cite{shan2018lego}) and SROM with and without dynamic speed removal at different speeds with the KITTI dataset and a real-time run in our campus (IIITH). (NOTE: LT implies lost tracking). It can be observed that we are always better than the performance of both the techniques at high-speed maneuvers. We are also able to register the map at $10Hz$.\normalsize}
\vspace{-0.4cm}
\label{tab:table1}
\end{table}
\end{center}
\normalsize



\section{Conclusion and future work}
In this paper, we presented SROM a simple real-time SLAM system for the autonomous vehicle. SROM gives a highly accurate 6-DOF transformation with a two-layer structure wherein the initial pose is estimated using Phase Only Correlation method followed by the point-to-plane ICP algorithm to obtain fine matching. Another key feature we proposed is the dynamic object removal which helps in achieving better performance and low drifts. We also showcased SROM’s ability to maintain localization at low sampling rates or at high linear or angular velocities where most popular LiDAR-based localization approaches get degraded fast. We exhaustively evaluated the proposed method in many challenging highways/country/urban sequences from the  KITTI  dataset and the results demonstrate better accuracy in comparisons to other state-of-the-art methods with reduced computational complexity aiding in real-time realizations. We have also integrated our SROM system with our in-house autonomous vehicle and compared it with the state-of-the-art methods like  LOAM  and  LeGO-LOAM. We plan to extend this work by exploring various sampling techniques for feature extraction from a point cloud. This can further reduce the computational complexity and accuracy of the proposed method. We also plan to use the occupancy maps for loop closure detection and performing a pose graph optimization on the poses obtained from consecutive LiDAR sweeps.
\vspace{-0.2cm}

\addtolength{\textheight}{-6cm}
\bibliographystyle{IEEEtran}
\bibliography{our}

\end{document}